%% file: main.tex
\documentclass[10pt, conference]{IEEEtran}
\IEEEoverridecommandlockouts
\usepackage{cite}
\usepackage{authblk}
\usepackage{marvosym}
\usepackage{textcomp}
\usepackage[a4paper]{geometry}

\def\BibTeX{{\rm B\kern-.05em{\sc i\kern-.025em b}\kern-.08em
    T\kern-.1667em\lower.7ex\hbox{E}\kern-.125emX}}
\input{config}

\begin{document}
\begin{CJK*}{UTF8}{gbsn}

\title{Adaptive Reconvergence-driven AIG Rewriting via Strategy Learning}

\author{ Liwei~Ni$^{1,2,6}$, Zonglin~Yang$^{3}$, Jiaxi~Zhang$^{4}$, Junfeng~Liu$^{5}$, Huawei~Li$^{1,2,6}$, Biwei~Xie$^{1,2,6}$ and Xinquan~Li$^{2,\text{\Letter}}$ \\
$^1$Institute of Computing Technology, Chinese Academy of Sciences, Beijing, China \\
$^2$Peng Cheng Laboratory, Shenzhen, China \\
$^3$Shenzhen University, Shenzhen, China \\
$^4$School of Computer Science, Peking University, Beijing, China \\
$^5$SKLSDE, Beihang University, Beijing, China \\
$^6$University of Chinese Academy of Sciences, Beijing, China \\
Emails: nlwmode@gmail.com, 2100271085@email.szu.edu.cn, zhangjiaxi@pku.edu.cn, \\ 
liujunfeng@buaa.edu.cn, lihuawei@ict.ac.cn, xiebiwei@ict.ac.cn, lixq01@pcl.ac.cn 
\thanks{This work is supported in part by the Major Key Project of PCL (No. PCL2023AS2-3) and NSFC (No. 62090024).}
}

\maketitle

\begin{abstract}
Rewriting is a common procedure in logic synthesis aimed at improving the performance, power, and area (PPA) of circuits.
The traditional reconvergence-driven And-Inverter Graph (AIG) rewriting method focuses solely on optimizing the reconvergence cone through Boolean algebra minimization. However, there exist opportunities to incorporate other node-rewriting algorithms that are better suited for specific cones.
In this paper, we propose an adaptive reconvergence-driven AIG rewriting algorithm that combines two key techniques: multi-strategy-based AIG rewriting and strategy learning-based algorithm selection.
The multi-strategy-based rewriting method expands upon the traditional approach by incorporating support for multi-node-rewriting algorithms, thus expanding the optimization space.
Additionally, the strategy learning-based algorithm selection method determines the most suitable node-rewriting algorithm for a given cone.
Experimental results demonstrate that our proposed method yields a significant average improvement of 5.567\% in size and 5.327\% in depth.
\end{abstract}


\section{introduction}
\label{sec:intro}
Logic level optimization~\cite{synthesis} is a crucial process in reducing the area and depth of a circuit to achieve better performance, power, and area~(PPA).
Rewriting, a highly flexible and efficient technique, is widely used for logic level optimization, providing a means to reduce circuit costs in many ways. The circuit is typically represented as a Directed Acyclic Graph~(DAG), and the general framework for rewriting can be summarized in the following four steps:
\begin{itemize}
    \item[1)] \textbf{Subgraph computation}.
    \item[2)] \textbf{Possible structures computation}.
    \item[3)] \textbf{Cost evaluation}.
    \item[4)] \textbf{Equivalent local replacement}.
\end{itemize}

Numerous works on rewriting have focused on the aforementioned steps, and the following provides a brief introduction:
Regarding the \textbf{\textit{Subgraph computation}} step, the emphasis is primarily on multi-inputs and single-output subgraphs. Works such as \cite{NPN-rewrite-4}, \cite{NPN-rewrite-5}, and \cite{exact-rw-date20} have primarily employed $k$-feasible cut enumeration \cite{priority-cuts} to compute subgraphs. However, due to the exponential growth of exhaustive cut enumeration with input size $k$, this method is typically used when $k$ is less than or equal to 4.
On the other hand, works such as \cite{brayton2006scalable}, \cite{reconvergent-window}, and \cite{exact-rw-date19} focus on computing the reconvergence cone to obtain large-inputs subgraphs.
Once the subgraph is obtained, the next step is \textbf{\textit{Possible structures computation}}, which involves determining the potential lower-cost structure candidates.
In \cite{NPN-class}, possible structures are computed by determining their NPN classes through Negating and Permuting inputs or Negating the output.
Similarly, \cite{isop} computes possible structures using the irredundant sum-of-product~(iSOP) technique based on Boolean algebra.
For minimal delay/area structures, \cite{exact-synthesis-TCAD20} employs exact synthesis through Boolean chain encoding and SAT solvers.
And \cite{sop-balance-iccad11} focuses on computing balanced structures using the AND-tree-balancing algorithm.
The next step is the crucial \textbf{\textit{Cost evaluation}}, which assesses the benefits of substituting the original structure with the new one.
The depth can be estimated by the positive slack \cite{slack}, while the area is estimated by maximum fanout free cone~(MFFC)\cite{cut_ranking}.
In the case of the Boolean matching problem \cite{boolmatching}, a transformation correspondence is established between the inputs and outputs of the original subgraph and the new structure. The \textbf{\textit{Equivalent local replacement}} step ensures the equivalence after substituting the old subgraph with the new one in the original circuit.
These four steps are executed for each node, typically applying the rewriting algorithm to all nodes in a topological order.

Since the problem of area and depth minimization in logic synthesis is NP-Complete, there are also some works to apply deep learning in logic synthesis.
It is clear that the rewriting algorithm follows greedy approaches based on the local search method, and it can be easily trapped into local minimal that does not allow improving quality of results~(QoR)\cite{NPN-rewrite-5}.
However, a better result can be obtained through the design space exploration~(DSE) method. 
Works such as \cite{DSE_1}\cite{DSE_2}\cite{DSE_3}\cite{DSE_4} focus on obtaining improved area or depth results by exploring the design space of logic optimization sequences through RL and Bayesian Optimization. 
Additionally, \cite{AISYM} and \cite{AIMAP} optimize circuit size and mapping results using RL techniques.

Regarding the above brief introduction, we have three findings: 1) In the \textbf{\textit{Possible structures computation}} step, the node-rewriting algorithms exhibit varying optimization capabilities; 2) Reconvergence-cone-based rewriting primarily optimizes using the iSOP method. However, not all cones are large, and other node-rewriting algorithms can be applied to suitable cones as well; 3) During each iteration of node-rewriting, the previously modified substructures can be overwritten and refreshed by the subsequent cones.

In this paper, we propose an adaptive reconvergence-driven AIG rewriting algorithm that addresses the aforementioned three cases by incorporating a strategy learning approach.
Firstly, we enhance the original reconvergence-driven AIG rewriting algorithm to accommodate multiple node-rewriting algorithms.
Secondly, as it can be challenging to determine the most suitable node-rewriting algorithm for a given subgraph, we formulate this as a classification problem that involves predicting the node-rewriting algorithm based on the subgraph's features. To accomplish this, we propose a reinforcement learning (RL) algorithm to generate the training dataset for the classification problem.
Our contributions can be summarized as follows:
\begin{itemize}
    \item We propose an adaptive reconvergence-driven AIG rewriting algorithm against the three identified findings.
    \item We introduce a strategy learning approach for selecting the most suitable node-rewriting algorithm for a given subgraph by classification method. 
    \item We formulate the multiple node-rewriting selection algorithm as the Markov decision process, and solve it by RL to generate the training dataset.
    \item We demonstrate the effectiveness of our proposed algorithm through experiments and comparisons with traditional methods.
\end{itemize}

The remaining sections of the paper are organized as follows:
Section~\ref{sec:prelim} presents the preliminary concepts and background related to the reconvergence-driven rewriting algorithm and reinforcement learning.
Section~\ref{sec:Problem} outlines the problem formulation of our proposed rewriting algorithm and provides a detailed description of the Markov decision process formulation.
Section~\ref{sec:Framework} presents the framework of our proposed method, explaining the key components and steps involved.
Section~\ref{sec:Eval} presents the evaluation of our implementation through comprehensive experimentation, providing sufficient experimental results and analysis.
Finally, section~\ref{sec:conclu} concludes this paper.

\section{preliminaries}
\label{sec:prelim}
\subsection{Boolean Network}
\label{sec:prelim-logic}
A $Boolean\ network$ is modeled as a DAG, where the node denotes a logic gate, input or output, and an edge denotes a signal wire between two logic gates.
A node $v$ has zero or more $fanins$ and $fanouts$, and fanin denotes the number of incoming edges, and fanout denotes the number of outgoing edges. 
The $primary\ inputs$~(PIs) are the nodes that have zero fanins, and the $primary\ outputs$~(POs) are the nodes that have zero fanouts, we can intuitively think of PIs and POs are the inputs and outs of the circuit. 
The node $v$ is $dead$ \textit{iff} it is an internal logic gate without any fanouts.
A $transitive\ fanin/fanout\ cone$~(TFI/TFO) of the node $v$ is a subset of nodes that are reachable through the fanin/fanout edges of $v$. 
If there is a path from node $v_i$ to $v_j$, then we can say $v_i$ is in the TFI of $v_j$, and $v_j$ is in the TFO of $v_i$.
A $max\ fanout\ free\ cone$~(MFFC) of the node $v$ is a subset of the TFI of $v$, such that each path from a node in this subset to the POs must pass through $v$. 
An And-Inverter Graph~(AIG) is a Boolean network with the internal nodes referring to AND gate and the edges referring to the inverter-embed signal or not.
\subsection{Reconvergence-cone}
\label{sec:prelim-rewriting}

\begin{figure}[t]
    \centering
    \includegraphics[scale=0.5]{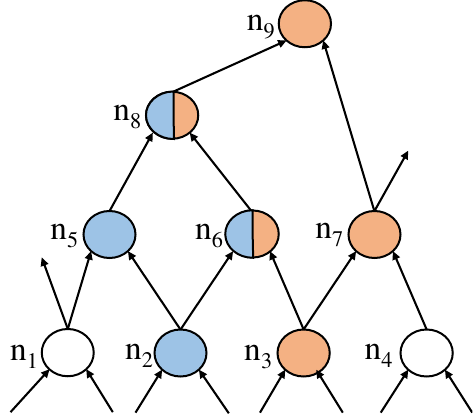}
    \caption{The subgraph structure with reconvergence in the Boolean network.}
    \label{fig:reconvergence_cone}
\end{figure}

The $reconvergence$ refers to the paths starting at the fanout of a node $n$ that will meet again before arriving at the POs. As shown in Fig. \ref{fig:reconvergence_cone}, it is a subgraph from a circuit, node $n_2$'s fanouts meet at $n_8$ before reaching POs, and as well as the path from $n_3$ to $n_9$.
The reconvergence cases are inevitable due to the logic sharing in multi-level Boolean networks.
The $reconvergence{-}cone$ technique works by identifying points in the circuit where multiple paths converge and then diverge again. 
The core idea is that there is a high probability of eliminable points on these paths in the reconvergence cone.
These points are known as $reconvergence\ points$, and they often represent opportunities for optimization.

\subsection{Reinforcement Learning}
\label{sec:prelim-RL}
RL is a subfield of Artificial Intelligence~(AI) that focuses on developing algorithms and techniques for an agent to learn optimal behaviors through interaction with an environment.
In RL, an agent learns by taking action in an environment and receiving feedback in the form of rewards or punishments.
The agent's goal is to learn a policy, which is a mapping from states to actions, that maximizes the cumulative rewards over time.

\subsubsection{Markov Decision Process~(MDP)}
\label{sec:prelim-RL-MDP}
An MDP is a mathematical framework used to model decision-making problems in which an agent interacts with a dynamic environment.
And it can be defined as the following:

\begin{equation}
\label{equ:mdp}
\textit{MDP} : (S, A, P, R, \gamma)
\end{equation}
where $S$ is the set of possible states~(state space), and $A$ is the set of possible actions~(action space).
$P$ is the state transition probability matrix, $P:S \times A \times S \rightarrow{\mathbb{R}_{\in{[0,1]}}}$, where $P(s, a, s')$ represents the probability of transitioning from state $s$ to state $s'$ when taking action $a$.
$R$ is the reward function, $R:S \times A \rightarrow{\mathbb{R}}$, where $R(s, a, s')$ represents the immediate reward corresponding to the $P(s, a, s')$.
And $\gamma$ is the discount factor.

The objective in an MDP is to find a policy $\pi(a|s)$, which is a mapping from states to actions, that maximizes the expected cumulative reward. 
This is typically done by defining a value function $V(s)$, which represents the expected cumulative reward starting from state s and following a particular policy π. The value function can be recursively defined using the Bellman equation:
\begin{equation}\
\label{eq:bellman}
V(s) = R(s, a, s') + \gamma\sum{ P(s, a, s')V(s')}
\end{equation}
By solving the equation (\ref{eq:opt_bellman}), the optimal value function $V^{*}(s)$ can be obtained, and the corresponding optimal policy $\pi^{*}(s)$ can be derived by selecting the action with the highest expected rewards in each state:
\begin{equation}
\label{eq:opt_bellman}
\pi^{*}(s) = \max_{a'}\{R(s, a', s') + \gamma\sum{ P(s, a', s')V(s')}\}
\end{equation}

\subsubsection{Q-learning}
\label{sec:prelim-RL-Q_learning}
Q-learning\cite{q-learning} is a popular RL algorithm that aims to learn an optimal policy for a Q-agent to make decisions in an environment.
In the Q-learning algorithm at any timestamp $t$, the Q-agent is at state $s_t$, it will enter the next state $s_{t+1}$ by taking action $a_t$. 
The value function is represented by the matrix of Q-value for each pair of state and action, $Q:S \times A \rightarrow\mathbb{R}$.
And the corresponding Bellman equation:
\begin{equation}\
\label{equ:Qbellman}
\begin{aligned}
Q(s_t, a_t) = &(1-\alpha)Q(s_{t-1}, a_{t-1}) + \\
              &\alpha(R(s_t, a_t) + \gamma \max_{a'}{Q(s_{t+1}, a')})
\end{aligned}
\end{equation}
where $\alpha \in{[0,1]}$ is the learning rate.
And there is also some proof in \cite{AISYM}\cite{q_learning_convergence} to tell that a finite MDP$(S, A, P, R, \gamma)$ can be solved optimally using Q-learning with the update rule as in equation (\ref{equ:Qbellman}) while the learning rate $\alpha$ is independent on both the state and action.

\section{Problem definition}
\label{sec:Problem}
In this section, we present the fundamental problem definition of our proposed multi-strategy rewriting algorithm. Furthermore, we conduct an analysis of the design space exploration for multi-strategy rewriting to demonstrate the significance of solving for the optimal strategy sequence using RL. Lastly, we formulate the aforementioned problem as a Markov decision process~(MDP).
\subsection{Multi-Strategy Selection Rewriting Problem}
\label{sec:Problem-Rewriting}

The traditional rewriting problem of the Boolean network, as introduced in Section~\ref{sec:intro}, performs node-rewriting at each node to reduce the cost of the circuit.
And it can be formulated as follows:

\begin{equation}
\label{equ:tra_rewriting}
\begin{aligned}
\text{min}~~&\sum_{i=0}^{m}{cost(n_i')-cost(n_i)} \\
\text{s.t.}~~&G' = { \underbrace{ \textit{NR}(~\cdot\cdot\cdot(~\textit{NR}(n_i)~)~) }_{\substack{m}}, ~~n_i \in {G}} \\
\end{aligned}
\end{equation}
where $G$ refers to the input Boolean network, and $G'$ refers to the optimized Boolean network and $n_i' \in G'$, $m$ is the node size of $G$, and the nodes generally follow a topological order.
The term \textit{NR} refers to a certain node-rewriting operation, which remains fixed once selected for all the nodes.
And the objective function is to minimize the cost of the circuit.

As also mentioned earlier in Section \ref{sec:intro}, our findings reveal that the reconvergence-driven rewriting algorithm has untapped optimization potential. To address this, we propose a multi-strategy selection rewriting algorithm, which can be formulated as follows:

\begin{equation}
\label{equ:str_rewriting}
\begin{aligned}
\text{min}~~&\sum_{i=0}^{m}{cost(n_i')-cost(n_i)} \\
\text{s.t.}~~&G' = {\underbrace{ \textit{NR}_k(\cdot\cdot\cdot(\textit{NR}_j(n_i))) }_{\substack{m}}, n_i \in {G},} \\
             &~~~~~~~~~~~~~~~~~~~~~~~~~~~{\textit{NR}_j,\cdots,\textit{NR}_k \in \textit{NRP}} \\
\end{aligned}
\end{equation}
where the $\textit{NR}_j$ and $\textit{NR}_k$ mean that we can use different node-rewriting algorithms at different nodes. the \textit{NRP} represents the available rewriting algorithm pool, and the remaining definitions and objectives remain the same as in equation (\ref{equ:tra_rewriting}).

\noindent\textbf{Opportunities:}
\begin{itemize}
    \item It offers a fusion approach that combines multiple node-rewriting methods;
    \item It has the potential to optimize the circuit further by leveraging the additional optimization space provided by different node-rewriting algorithms;
\end{itemize}
\noindent\textbf{Challenges}: 
\begin{itemize}
    \item How to decide the most suitable node-rewriting algorithm for a given cone?
    \item How to improve the generalization ability for the decision-making process for different AIG and reconvergence cones?
\end{itemize}

\subsection{Design Space Exploration of the Rewriting}
\label{sec:Problem-DSE}

\begin{figure}[t]
    \centering
    \includegraphics[scale=0.5]{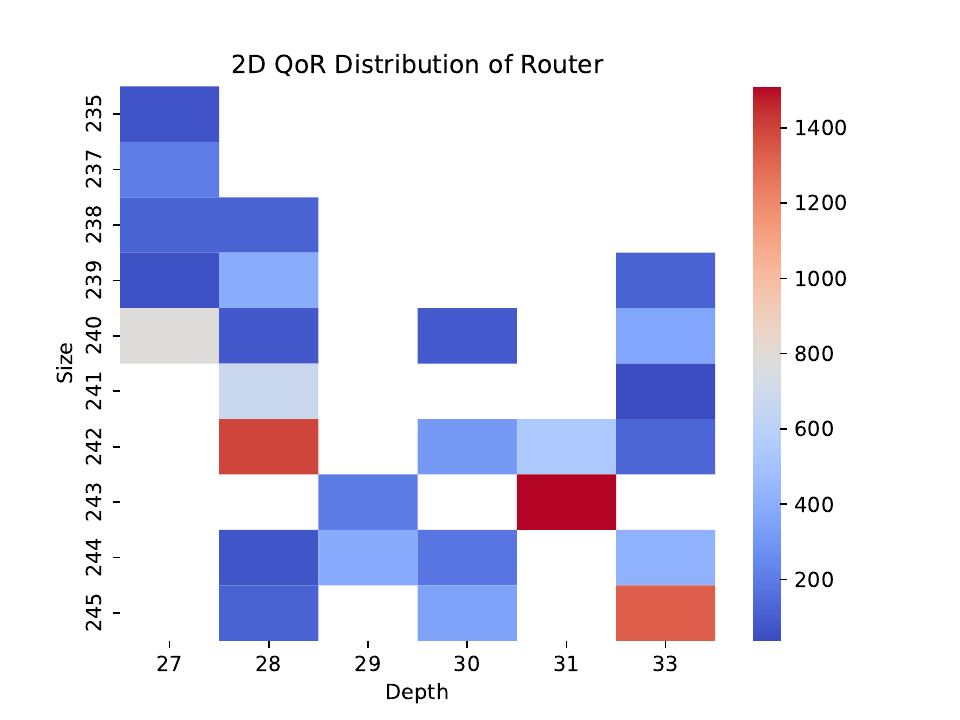}
    \caption{The 2D QoR Distribution for the case ``router".}
    \label{fig:dse}
\end{figure}

\begin{figure*}[t]
    \centering
    \includegraphics[scale=0.49]{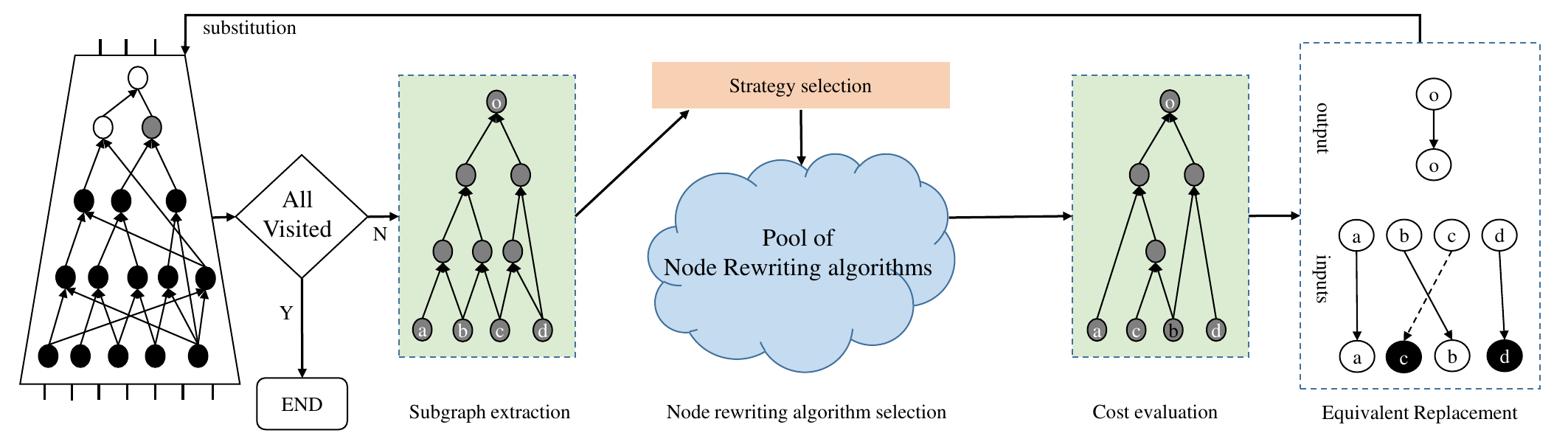}
    \caption{The framework of the multi-strategies selection-driven node-rewriting method. These steps are corresponding to the traditional rewriting method: (1) Cone(subgraph) extraction; (2) Possible structures computation; (3) Cost evaluation; (4) Equivalent replacement. The main difference here is the multi-strategies selection at step (2) corresponding to the equation~(\ref{equ:str_rewriting}) in subsection~\ref{sec:Problem-Rewriting}.}
    \label{fig:framework}
\end{figure*}
To demonstrate the effects of the multi-strategy rewriting method, we implemented a reconvergence-driven AIG rewriting algorithm using random strategies. 
These random strategies involve randomly selecting a node-rewriting algorithm from the algorithm pool for each rewritable node, taking into account the specific input size limitations of certain algorithms.

Using this novel optimization operator, we conducted an exploration of the design space by randomly shuffling the node-rewriting algorithms at each node. 
Specifically, we generated 10,000 rewriting results for the ``router'' case in the EPFL\cite{benchmark_epfl} benchmark using our reconvergence-driven AIG rewriting method based on random strategies.

The 2-D QoR distribution is presented in Fig. \ref{fig:dse} as a heatmap. 
The x-axis represents the depth of the resulting AIG, the y-axis represents the size, and the color indicates the count size. 
It is evident from the heatmap that the quality of results (QoR) is influenced by the node-rewriting algorithms, highlighting the importance of algorithm selection in achieving desired outcomes.

\subsection{Formulating Rewriting Selection as MDP}
\label{sec:Problem-MDP_From}

As mentioned in \ref{sec:Problem-Rewriting}, it tells that the node-rewriting algorithm is hard to be selected for the corresponding subgraph structure for the proposed multi-strategies-based problem.
In this section, we model the multi-strategies selection rewriting-based optimization into an MDP. 
It should also be noticed that the outcome of a deterministic strategy sequence for the nodes is also deterministic, which simplifies the common probabilistic setting in typical MDP formulations.
And we formulate this MDP problem in the rest of this subsection.

\subsubsection{State and its Space $S$} As the next node's rewriting follows the result circuit graph of the previous node's rewriting, we define the state as the extracted features of the computed reconvergence cone, and they are as follows:
\begin{itemize}
    \item $is\_critical$. It means the node is on the critical path or not, its range is Boolean\{0,1\};
    \item $input\_size$. It means the input size of the cone, and is sensitive to the node-rewriting algorithm, its range is [2,10];
    \item $node\_size$. It is the total size of this cone, the more node size with less input\_size may have more optimization space, its range is [2,16];
    \item $fanout\_size$. The total fanout size of the nodes in this cone. The MFFC is related to the fanout size;
    \item $positive\_edges$. It means the number of edges in two AND nodes without NOT gate;
    \item $negative\_edges$. It means the number of edges in two AND nodes with NOT gate;
    \item $max\_depth$. The max difference in depth between the root and leaves.
\end{itemize}
According to equation~(6), we can make a simple assumption that each node-rewriting algorithm will output different results/states for a cone, and the state space is $k^m$ in theory, where $k$ is the number of node-rewriting algorithms, $m$ is the number of times for node-rewriting which approximately equal to the number of internal nodes in AIG. In this paper, the state space's size is $3^m$ in theory.

\subsubsection{Action Space $A$} The action space is the pool of the following three node-rewriting algorithms. Since different node-rewriting algorithms have their own suitable input ranges, we also set some restrictions on them.
\begin{itemize}
    \item NPN-based node-rewriting\cite{NPN-rewrite-4}. As the search space for the NPN problem is $n! 2^{n+1}$($n$ is the input size), it is suitable when $n \in [2,4]$;
    \item Exact synthesis-based node-rewriting\cite{exact-synthesis-TCAD20}. Exact synthesis can compute the optimal area/depth for the given cone with the price of time overhead. Cache-based exact synthesis\cite{exact-rw-date20} can significantly reduce the time overhead, and we set the input size $n$ with the range of $[2,5]$.
    \item iSOP-based node-rewriting\cite{isop}. The iSOP method is the node-rewriting algorithm in the open source Logic Synthesis tool berkeley-abc\cite{berkeley-abc} as the command ``refactor'', and we set the input size $n$ with the same range of $[2,10]$.
\end{itemize}
\subsubsection{Rewards $R$} The reward function is directly corresponding to the QoR improvement, and it includes two parts:
\begin{itemize}
    \item $local\_reward$. It is the reward corresponding to the local gain after taking an action, and the local gain can be set to area, depth, or and-delay-product~(ADP).
    \item $global\_reward$. It is the total local reward after the train of one episode, and we set it by the total area gain here.
\end{itemize}

\section{Proposed framework}
\label{sec:Framework}
In this section, we provide the framework of our proposed adaptive reconvergence-driven AIG rewriting method via the strategy learning method.
The core idea is to support the multi-node-rewriting algorithm for a given cone.
And the implementation scheme is to classify the features we defined by a classifier, before that, the labeled dataset is generated by a Q-learning procedure. 

\subsection{Framework Overview}
\label{sec:Framework-Rewriting}
\vspace{5pt}
\begin{algorithm}[ht]
\small
\caption{Reconvergence-driven AIG rewriting via Multi-Strategy Selection}
\label{algorithm:flow}
\begin{algorithmic}[1]
\Require Original AIG $G$
\Ensure Optimized AIG $G'$
\State $V$ $\leftarrow$ topological\_sorting($G$)
\State critical\_path\_computation($G$)
\For{$n$ in $V$}
    \If{is\_logic\_node($n$)}
        \State $sg$ $\leftarrow$  reconvergence\_cone\_computation($n$)
        \State $fe$ $\leftarrow$  feature\_extraction($sg$)
        \State $op$ $\leftarrow$  \textbf{strategy\_selection}($fe$)
        \State $psg$ $\leftarrow$ possible\_structure\_computation($op$, $sg$)
        \State $price$ $\leftarrow$ cost\_evaluation($sg$, $psg$)
        \If{$price$\ is\ acceptable:}
            \State $mp$ $\leftarrow$  match\_ports($sg$, $psg$)
            \State equivalent\_replacement($mp$, $sg$, $psg$)
        \EndIf
    \EndIf
\EndFor
\State $G'$ $\leftarrow$ remove\_dead\_nodes(G)
\end{algorithmic}
\end{algorithm}

The framework depicted in Figure \ref{fig:framework} presents an overview of our proposed method. 
The input Boolean network is situated on the left side of the framework. 
It is processing at the gray node, subsequently, we apply the multi-node-rewriting selection algorithm to the given subgraph, and the specific details of this process will be elaborated upon in Algorithm \ref{algorithm:flow}.

As depicted in Algorithm \ref{algorithm:flow},
the rewriting algorithm follows a topological order~(line 1), and the critical path also should be marked for the influence of depth~(line 2).
Then we perform the multi-node-rewriting algorithm at each logic node~(lines 3-15).
For a given processing node $n$, the reconvergence cone $sg$ will be computed first~(line 5);
Following this, the previously defined features $fe$ are extracted to enable the evaluator to determine the appropriate node-rewriting algorithm $op$ to be selected~(lines 6-7). 
The possible structure candidates $psg$ will be computed by $op$, and the cost $price$ also will be evaluated whether it will result in the QoR improvement~(lines 8-9).
If the $price$ is deemed acceptable~(positive gains), the ``equivalent local replacement'' is executed to insert $psg$ into the original circuit by ensuring correct port matching (lines 11-12).

We highlight two key aspects in our framework:
(1) The ``multi-strategy-based'' rewriting method can expand the search space of possible structures by the algorithm pool;
(2) The ``strategy selection'' process can be implemented using a learning approach.
In the following two subsections, we will elaborate on how RL is employed to generate the training dataset and how the classification method is employed for ``strategy selection''.

\subsection{Training dataset generation through Q-learning}
\label{sec:Framework-RL}

\vspace{5pt}
\begin{algorithm}[ht]
\small
\caption{Q-learning based training procedure}
\label{algorithm:q-learning}
\begin{algorithmic}[1]
\Require Input AIG G
\Ensure Policy matrix: Q-value table
\State Init Q-learning agent: $q\_agent$ with $\gamma,\ \alpha$
\State Init train parameter: $st$, $Eps$, $decay$, $\epsilon$, $done$ $\leftarrow$ false
\For{$ep$ in range($Eps$)}
    \While{not $done$}
        \State $st$ $\leftarrow$ compute\_current\_state($G$)
        \State $pas$ $\leftarrow$ get\_possible\_actions($st$)
        \State $a$ $\leftarrow$ choose\_action($q\_agent$, $st$, $pas$)
        \State $nst, r$ $\leftarrow$ apply($st$, $a$, $\epsilon$)
        \If{$nst$ is valid}
            \State update\_q\_table($q\_agent$, $a$, $r$)
            \State $st$ $\leftarrow$ $nst$
        \Else
            \State $done$ $\leftarrow$ true
        \EndIf
        \State $\epsilon$ $\leftarrow$ $\epsilon$ * $decay$
    \EndWhile
\EndFor
\end{algorithmic}
\end{algorithm}

We formulate the optimal ``strategy selection'' problem as an MDP as mentioned in subsection~\ref{sec:Problem-MDP_From}.
In this subsection, we use the Q-learning\cite{q-learning} method as the ``agent'' to learn a good policy of the circuits.
The following provides a detailed description of the implementation:

\subsubsection{Strategy sequence} 

\begin{figure}[t]
    \centering
    \includegraphics[scale=0.7]{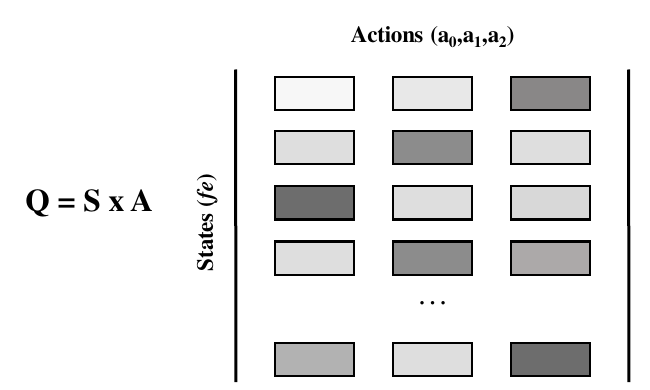}
    \caption{The Q-table for the MDP solving by the Q-learning method.}
    \label{fig:q-learning}
\end{figure}

To facilitate the calculation of the current state and next state during the topological node-rewriting procedure, we introduce the ``strategy sequence'' method, which fixes the strategy at each rewritable node.
First, we create a mapping for each node-rewriting algorithm as follows: \{\{``iSOP'', 0\}, \{``Exact synthesis'', 1\}, \{``NPN'', 2\}\}. 
Each algorithm is assigned a value that corresponds to the index of the Actions \{$a_0, a_1, a_2$\}.
The following Example \ref{exm:sequence} illustrates how the strategy sequence helps in computing the current state and next state.

\begin{example}
\label{exm:sequence}
For an AIG network under the given strategy sequence ``01002011'', the first eight rewritable nodes will use the corresponding node-rewriting algorithm to optimize its reconvergence cone.
And the current state is the features of the 8th node's reconvergence cone, the next state will be computed by the 9th~(8+1$\rightarrow$9) computed reconvergence cone.
\end{example}

\subsubsection{Q-learning-based policy learning}

Fig. \ref{fig:q-learning} displays the Q-table used in our method, which illustrates the process of updating the rewards for the defined states and actions.
Each state is represented by a tuple of the features $fe$, and the action space consists of three node-rewriting actions. 
Each rectangular node in the Q-table represents the reward $Q(s_t, a_t)$ for a state under its corresponding action. The brightness of the rectangular node indicates the value of the reward, and the darker the brightness, the greater the value.

Algorithm \ref{algorithm:q-learning} outlines the training procedure based on Q-learning.
The $q\_agent$ is initialized with the learning rate $\alpha$ and discount factor $\gamma$ according to the equation (\ref{equ:Qbellman})~(line 1).
We also initialize several training parameters: $st$ represents the current state, $Eps$ denotes the batch size of the training, $decay$ and $\epsilon$ indicate the discount factor when selecting actions~(line 2).
For each training episode, the Q-table is updated according to the current state $st$, its possible actions $pas$, and the computed reward $r$, and then proceeds to the next state $nst$~(lines 5-15).
It should be noted that the possible actions $pas$ are selected based on the restrictions set in subsection~\ref{sec:Problem-MDP_From}~(line 6).
If the next state is not valid, indicating that all rewritable nodes have been visited and rewritten, the procedure proceeds to the next episode~(line 13).

\subsection{Classification}
\label{sec:Framework-classification}
\begin{figure}[t]
    \centering
    \includegraphics[scale=0.48]{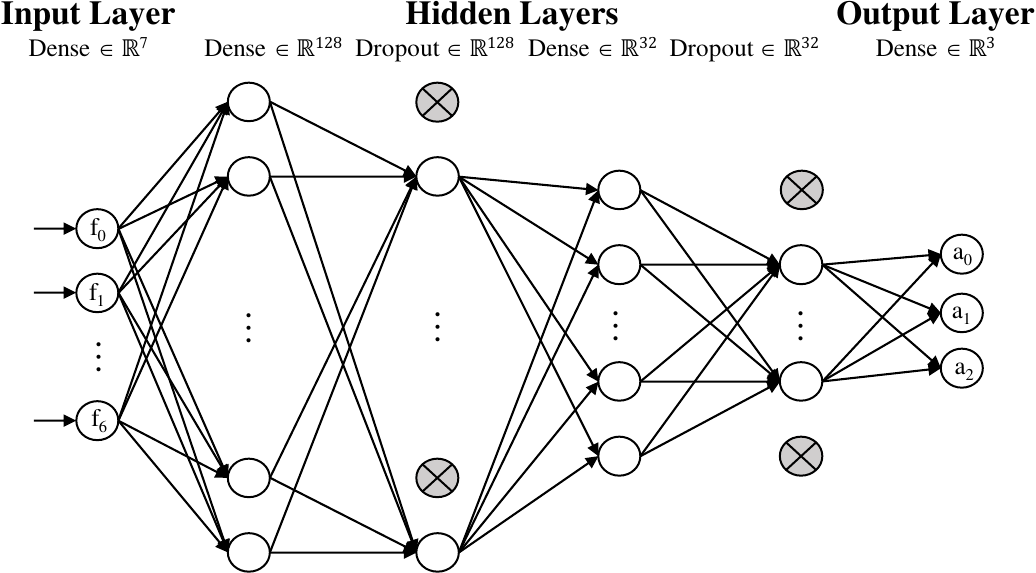}
    \caption{The architecture of MLP for classification.}
    \label{fig:classify-framework}
\end{figure}
Due to the dimension of the features is only 7, and the Q-learning procedure will provide the data of the best action for its corresponding state, also as the feature, all of this undoubtedly motivates us to use classification methods to solve the problem of generalization.
Hence, we employ a multi-layer Perceptron~(MLP)\cite{MLP} with dropout as a classification method to classify the features with the corresponding node-rewriting algorithm labels.

The architecture of the MLP is depicted in Fig. \ref{fig:classify-framework}, consisting of the input layer, hidden layers, and output layer.
The input layer has 7 dimensions, corresponding to the 7 features.
Following the input layer, there are two hidden layers with dimensions of 128 and 32, respectively, each followed by a dropout layer.
The dropout layer helps enhance the generalization ability of the neural network and mitigates overfitting.
Finally, the output layer of the MLP is fully connected to the hidden layer, and a softmax activation function is applied to obtain the probabilities of the output classes.

\section{evaluation}
\label{sec:Eval}
The implementation of our proposed method ``refactor-plus'' is using the open source Boolean network library ``mockturtle''\cite{EPFLLibraries} to reproduce the reconvergence-driven AIG rewriting method, also as the command ``refactor'', of the berkeley-abc project in C++ language.
Then we make it supports the multi-node-rewriting algorithm.
And the Q-learning and MLP classification codes are implemented in Python language.
All procedures run on an Intel(R) Xeon(R) Platinum 8260 CPU with 2.40GHz, 24 cores, and 128GB RAM.
As for the benchmark, we perform the evaluation on the well-known EPFL combinational\cite{benchmark_epfl} benchmark, which consists of 10 arithmetic circuits and 10 control circuits, and the circuit sizes range from 174 to 214335.

\subsection{Convergence of Strategy Learning}
\label{sec:Eval-convergence}
For the convergence of strategy learning, there are two aspects to consider: the convergence of Q-learning for the training dataset and the convergence of the classification model for the features.
\subsubsection{Convergence of Q-learning}
\begin{figure}[t]
\centering
\subfigure[ctrl]{
    \label{fig:a}
    \includegraphics[width=0.45\linewidth]{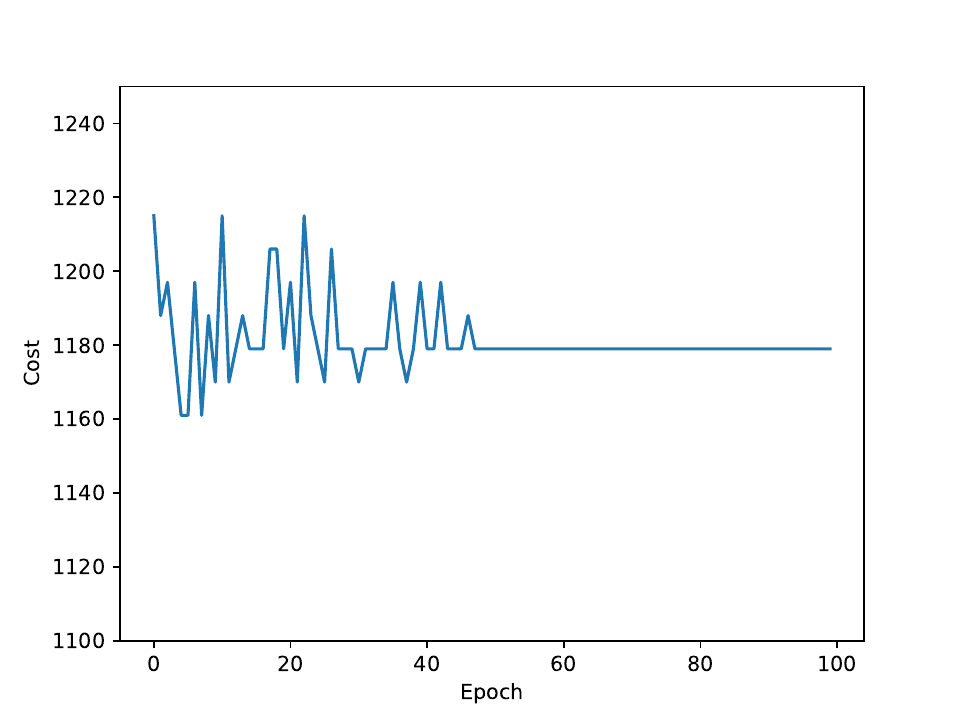}
    }
\subfigure[router]{
    \label{fig:b}
    \includegraphics[width=0.45\linewidth]{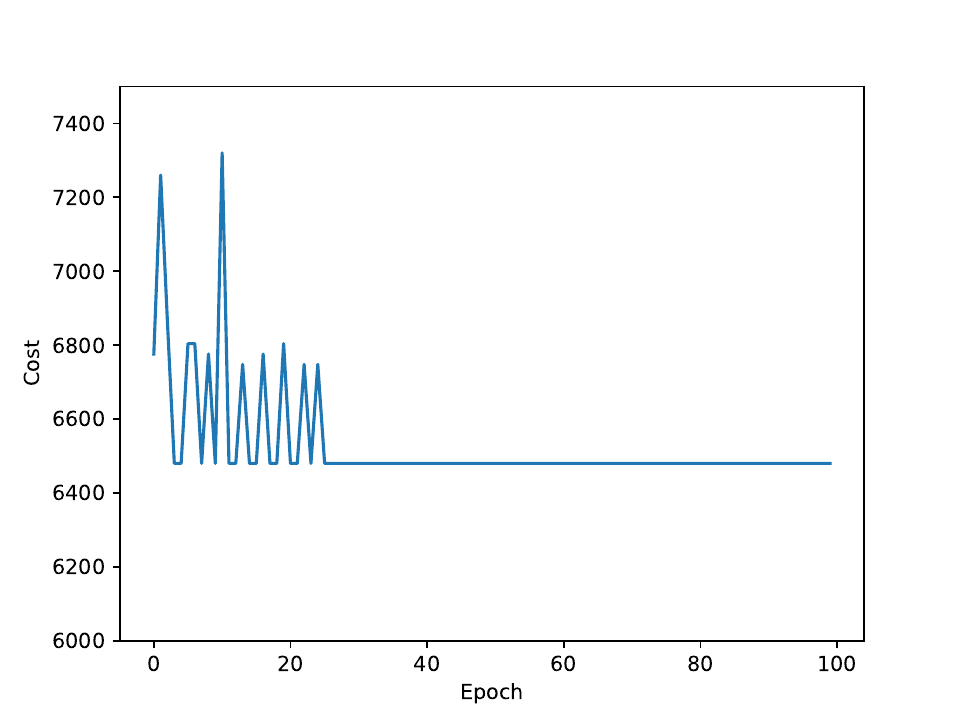}
    }
\subfigure[int2float]{
    \label{fig:c}
    \includegraphics[width=0.45\linewidth]{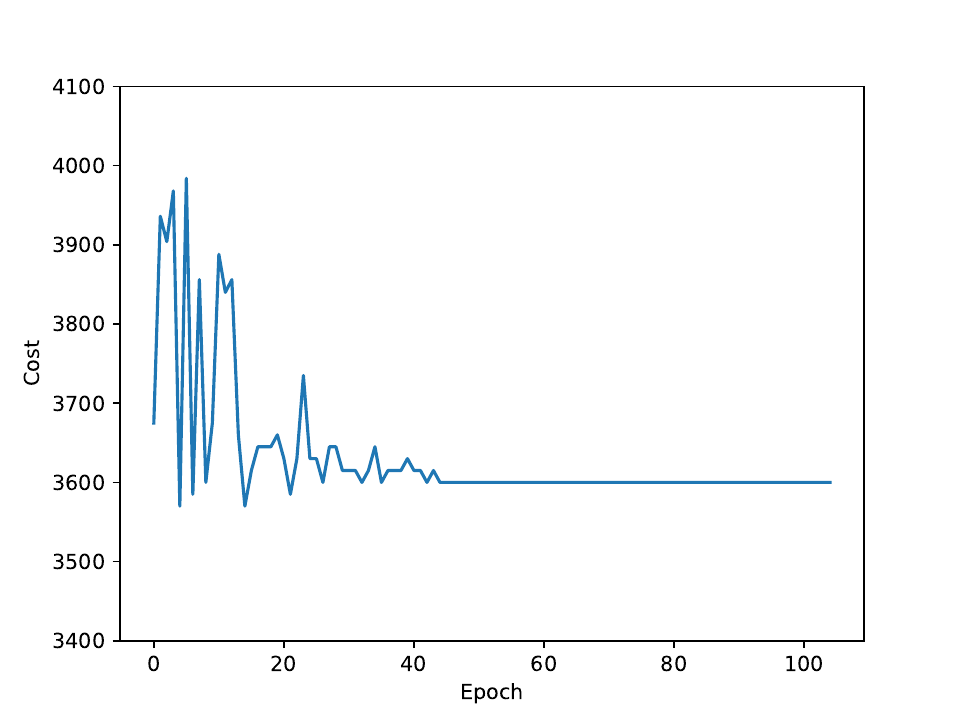}
    }
\subfigure[priority]{
    \label{fig:d}
    \includegraphics[width=0.45\linewidth]{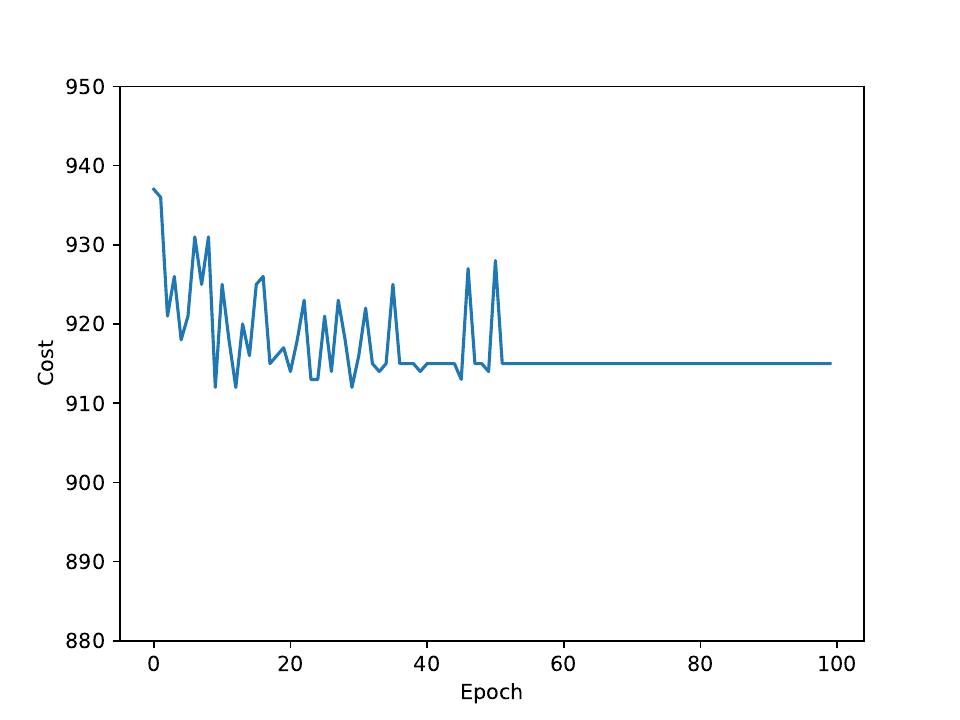}
    }
\caption{The convergence of the Q-learning strategy selection method, and the y-axis represents the cost value which is negatively related to the reward value of the learning procedure, and the smaller, the better. }
\label{fig:convergence-q-learning}
\end{figure}

Fig. \ref{fig:convergence-q-learning} demonstrates the convergence curve of the Q-learning training procedure over 100 episodes on 4 combinational circuits.
The curve indicates that the Q-learning algorithm achieves convergence within 50 episodes, it suggests that the training process is effective in learning the optimal policy for selecting node-rewriting algorithms.
Along with the Q-learning procedure converging, the classification procedure can utilize the well-trained dataset generated by Q-learning.
Because the RL-based training dataset generation procedure is time-consuming, here we did not give the program running time, but the reader can according to the TABLE-\ref{table:single-opt} and SECTION-\ref{sec:Framework-RL} to estimate the runtime roughly.

\subsubsection{Convergence of classification}
\begin{figure}[t]
    \centering
    \includegraphics[scale=0.5]{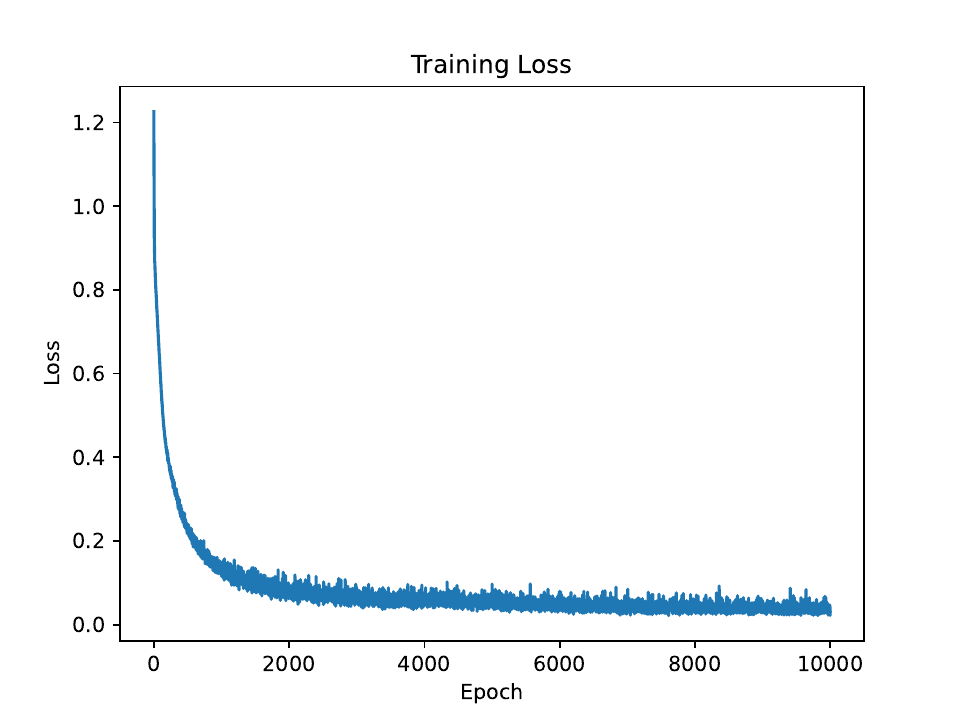}
    \caption{Train Loss of the classification problem on the features through MLP method with 87\% test accuracy.}
    \label{fig:convergence-classification}
\end{figure}
Fig. \ref{fig:convergence-classification} depicts the convergence of the classification procedure of features.
All the trained dataset is obtained from the Q-learning step on the small cases.
Since this is a multi-classification problem, the loss function employed is the ``cross-entropy loss''. 
The dataset is split into an 80\% training set and a 20\% test set for model training and evaluation, respectively.
After 10,000 training episodes, it shows that the loss function is converging. 
And the overhead for inference time is low, and it is about 10 microseconds according to the statistical results.
Furthermore, the test accuracy of 87\% suggests that the model is able to generalize well to the unseen cone's feature.
The following experiments will demonstrate the effectiveness of our proposed algorithm.

\subsection{Comparison with traditional ``refactor''}
\label{sec:Eval-Operator}

In this subsection, the experiments will depict that the multi-strategy selection rewriting through strategy learning is working well.
And our solution for strategy learning is also successful.
To showcase the performance of our proposed method ``refactor-plus'' compared to the traditional ``refactor'' method, we conducted experiments from the following four perspectives:

\subsubsection{Comparison by individual operator}
The performance of individual operator optimizations is presented in Table \ref{table:single-opt}.
The results demonstrate that the ``refactor'' method achieves an average improvement\footnote{For improv\% computation, Table \ref{table:single-opt} is computed by (origin - refactor/refactor-plus)/origin*100; Table \ref{table:post-balance}, \ref{table:resyn2}, and \ref{table:post-mapping} are computed by (refactor - refactor-plus)/refactor*100.} of 3.211\% in size and 1.482\% in depth.
While the ``refactor-plus'' method achieves a higher improvement of 5.567\% in size and 5.327\% in depth.
In most cases, ``refactor-plus'' outperforms ``refactor'' and effectively reduces both the size and depth of the circuits.
In the best case scenario, the improvement of ''refactor-plus'' can reach 24.138\% in size and 50\% in depth.
It also shows that ``refactor-plus'' can effectively reduce the depth while reducing the size than ``refactor''.
The time of ``refactor-plus'' is mainly caused by the three parts: (1)the selected algorithm's time; (2) the data structure we implemented; and (3) the decision inference time. As the inference is low overhead, the main causes are (1) and (2).

\subsubsection{Comparison by ``pre-balance''}

\begin{table}[t]
    \centering
    \caption{Comparison of the results on ``refactor'' and ``refactor-plus''.}
    \label{table:single-opt}
    \setlength{\tabcolsep}{1pt}
    \resizebox{\linewidth}{!}{
    \begin{tabular}{c|cc|ccccc|ccccc}
    \toprule
    \multicolumn{1}{c|}{\multirow{3}{*}{Circuits}} & \multicolumn{2}{c|}{\multirow{2}{*}{origin}} & \multicolumn{5}{c|}{refactor}                                                                 & \multicolumn{5}{c}{refactor-plus}           \\
    \multicolumn{1}{c|}{}                          & size     &  \multicolumn{1}{c|}{depth}       & size             &  improv\%     &depth         &improv\%    &  \multicolumn{1}{c|}{time(s) } & size                  &  improv\%        & depth          &  improv\%        & time(s)    \\ \hline \hline
    log2                                           & 32060    &     444                           & 31521            &  1.681        &   444        &   0        &  2.536                         & \textbf{30814}        &  3.886           &  \textbf{423}  &  4.730           & 9.677           \\
    square                                         & 18484    &     250                           & 18302            &  0.985        &   250        &   0        &  1.38                          & \textbf{18275}        &  1.131           &  \textbf{248}  &  0.800           & 2.051           \\
    adder                                          & 1020     &      255                          & 1019             &  0.098        &    255       &   0        &  0.066                         & 1019                  &  0.098           &  255           &  0               & 0.486           \\
    sin                                            & 5416     &      225                          & 5321             &  1.754        &    224       &   0.444    &  0.286                         & \textbf{5279}         &  2.530           &  \textbf{214}  &  4.889           & 0.95           \\
    div                                            & 57247    &     4372                          & 56745            &  0.877        &   4372       &   0        &  2.718                         & \textbf{56550}        &  1.218           &  4372          &  0               & 23.265           \\
    hyp                                            & 214335   &    24801                          & \textbf{212341}  &  0.930        &  24801       &   0        &  23.872                        & 212431                &  0.888           &  \textbf{24794}&  0.028           & 301.767           \\
    max                                            & 2865     &      287                          & 2865             &  0            &    287       &   0        &  0.154                         & 2865                  &  0               &  \textbf{226}  &  21.254          & 0.629           \\
    sqrt                                           & 24618    &     5058                          & 23685            &  3.790        &   5058       &   0        &  0.379                         & \textbf{20155}        &  18.129          &  \textbf{5056} &  0.040           & 8.107           \\
    multiplier                                     & 27062    &     274                           & 26814            &  0.916        &   274        &   0        &  2.171                         & \textbf{25509}        &  5.739           &  \textbf{272}  &  0.730           & 6.112           \\
    bar                                            & 3336     &      12                           & 3141             &  5.845        &    12        &   0        &  0.144                         & 3141                  &  5.845           &  12            &  0               & 0.545           \\
    priority                                       & 978      &       250                         & \textbf{854}     &  12.679       & \textbf{227} &   9.200    &  0.089                         & 888                   &  9.202           &  242           &  3.200           & 0.489           \\
    cavlc                                          & 693      &       16                          & 690              &  0.433        &     16       &   0        &  0.116                         & \textbf{688}          &  0.722           &  16            &  0               & 0.517           \\
    arbiter                                        & 11839    &     87                            & 11839            &  0            &   87         &   0        &  0.42                          & 11839                 &  0               &  87            &  0               & 2.886           \\
    i2c                                            & 1342     &      20                           & 1338             &  0.298        &    20        &   0        &  0.09                          & \textbf{1308}         &  2.534           &  \textbf{18}   &  10              & 0.49           \\
    voter                                          & 13758    &     70                            & 12681            &  7.828        &   63         &   10       &  0.677                         & \textbf{11363}        &  17.408          &  70            &  0               & 1.455           \\
    int2float                                      & 260      &       16                          & 251              &  3.462        &     16       &   0        &  0.065                         & \textbf{237}          &  8.846           &  16            &  0               & 0.485           \\
    ctrl                                           & 174      &       10                          & 143              &  17.816       &     9        &   10       &  0.065                         & \textbf{132}          &  24.138          &  \textbf{9}    &  10              & 0.394           \\
    dec                                            & 304      &       3                           & 304              &  0            &     3        &   0        &  0.063                         & 304                   &  0               &  3             &  0               & 0.345           \\
    mem\_ctrl                                      & 46836    &     114                           & 46582            &  0.542        &   114        &   0        &  0.875                         & \textbf{46071}        &  1.633           &  \textbf{113}  &  0.877           & 13.729           \\
    router                                         & 257      &       54                          & 246              &  4.280        &     54       &   0        &  0.075                         & \textbf{238}          &  7.393           &  \textbf{27}   &  50              & 0.455           \\  \hline \hline
    \textbf{AVE.}                                  &  -       & \multicolumn{1}{c|}{-}            &  -               &  3.211        &   -          &   1.482    &   -                            &  -           &  \textbf{5.567}           &    -           &  \textbf{5.327}  &   -        \\  \bottomrule
    \end{tabular}
    }
\end{table}

\begin{table}[t]
    \centering
    \caption{Comparison of the results of ``pre-balance''.}
    \label{table:post-balance}
    \resizebox{\linewidth}{!}{
    \begin{tabular}{c|cc|cccc}
    \toprule
    \multicolumn{1}{c|}{\multirow{3}{*}{Circuits}} & \multicolumn{2}{c|}{refactor+(balance)}     & \multicolumn{4}{c}{refactor-plus+(balance)}             \\
    \multicolumn{1}{c|}{}                          & size        & \multicolumn{1}{c|}{depth } & size         &  improv\%    & depth       &  improv\% \\ \hline \hline
    log2                                           & 31400       &  410                        & \textbf{30706}        &  2.210       &  \textbf{409}        &  0.244    \\
    square                                         & 18086       &  250                        & \textbf{18067}        &  0.105       &  \textbf{248}        &  0.800    \\
    adder                                          & 1019        &   255                       & 1019                  &  0           &   255       &  0        \\
    sin                                            & 5297        &   186                       & \textbf{5258}         &  0.736       &   \textbf{184}       &  1.075    \\
    div                                            & 56729       &  4372                       & \textbf{56533}        &  0.346       &  4372       &  0        \\
    hyp                                            & \textbf{211669}      & 24801              & 212115                &  -0.211      & \textbf{24792}       &  0.036    \\
    max                                            & 2865        &   287                       & 2865                  &  0           &   \textbf{200}       &  30.314   \\
    sqrt                                           & 23681       &  5058                       & \textbf{20139}        &  14.957      &  \textbf{5056}       &  0.040    \\
    multiplier                                     & 26705       &  266                        & \textbf{25399}        &  4.890       &  \textbf{264}        &  0.752    \\
    bar                                            & 3141        &   12                        & 3141                  &  0           &   12        &  0        \\
    priority                                       & \textbf{768}         &    \textbf{227}             & 805                   &  -4.818      &    241      &  -6.167   \\
    cavlc                                          & 686         &    16                       & \textbf{684}          &  0.292       &    16       &  0        \\
    arbiter                                        & 11839       &  87                         & 11839                 &  0           &  87         &  0        \\
    i2c                                            & 1275        &   16                        & \textbf{1247}         &  2.196       &   16        &  0        \\
    voter                                          & 12474       &  \textbf{62}                         & \textbf{11244}        &  9.861       &  70         &  -12.903  \\
    int2float                                      & 233         &    15                       & \textbf{220}          &  5.579       &    15       &  0        \\
    ctrl                                           & 142         &    9                        & \textbf{130}          &  8.451       &    9        &  0        \\
    dec                                            & 304         &    3                        & 304                   &  0           &    3        &  0        \\
    mem\_ctrl                                      & 46549       &  114                        & \textbf{45953}        &  1.280       &  \textbf{113}        &  0.877    \\
    router                                         & 246         &    27                       & \textbf{239}          &  2.846       &    \textbf{26}       &  3.704    \\  \hline \hline
    \textbf{AVE.}                                  &  -          &   -                         &  -                    &  \textbf{2.436}       &    -        &  \textbf{0.939 }   \\  \bottomrule
    \end{tabular}
    }
\end{table}

\begin{table}[t]
    \centering
    \caption{Comparison of the results of ``resyn2''.}
    \label{table:resyn2}
    \resizebox{\linewidth}{!}{
    \begin{tabular}{c|cc|cccc}
    \toprule
    \multicolumn{1}{c|}{\multirow{3}{*}{Circuits}} & \multicolumn{2}{c|}{resyn2 with refactor}     & \multicolumn{4}{c}{resyn2 with refactor-plus}        \\
    \multicolumn{1}{c|}{}                          & size        & \multicolumn{1}{c|}{depth } & size         &  improv\%    & depth       &  improv\% \\ \hline \hline
    log2                                           & 29370       & 376                         & \textbf{29257}        &  0.385       & \textbf{368}         & 2.128     \\                 
    square                                         & \textbf{16623}       & 248                         & 16636        &  -0.078      & \textbf{247}         & 0.403     \\                 
    adder                                          & 1019        & 255                         & 1019         &  0           & 255         & 0         \\                 
    sin                                            & 5039        & 177                         & \textbf{5034}         &  0.099       & \textbf{172}         & 2.825     \\                 
    div                                            & 40772       & \textbf{4361}                        & \textbf{40758}        &  0.034       & 4362        & -0.023    \\                 
    hyp                                            & \textbf{211330}      & 24794                       & 211408       &  -0.037      & \textbf{24786}       & 0.032     \\                 
    max                                            & \textbf{2834}        & 204                         & 2858         &  -0.847      & \textbf{179}         & 12.255    \\                 
    sqrt                                           & 19437       & \textbf{4968}                        & 19437        &  0           & 4975        & -0.141    \\                 
    multiplier                                     & 24556       & 262                         & \textbf{24379}        &  0.721       & 262         & 0         \\                 
    bar                                            & 3141        & 12                          & 3141         &  0           & 12          & 0         \\                 
    priority                                       & \textbf{676}         & \textbf{203}                         & 707          &  -4.586      & 217         & -6.897    \\                 
    cavlc                                          & 662         & 16                          & \textbf{656}          &  0.906       & 16          & 0         \\                 
    arbiter                                        & 11839       & 87                          & 11839        &  0           & 87          & 0         \\                 
    i2c                                            & 1162        & 15                          & \textbf{1143}         &  1.635       & \textbf{14}          & 6.667     \\                 
    voter                                          & 9756        & 57                          & \textbf{8798}         &  9.820       & 57          & 0         \\                 
    int2float                                      & 214         & 15                          & \textbf{210}          &  1.869       & 15          & 0         \\                 
    ctrl                                           & 108         & 8                           & \textbf{107}          &  0.926       & 8           & 0         \\                 
    dec                                            & 304         & 3                           & 304          &  0           & 3           & 0         \\                 
    mem\_ctrl                                      & 45614       & 110                         & \textbf{44949}        &  1.458       & \textbf{107}         & 2.727     \\                 
    router                                         & \textbf{177}         & \textbf{19}                          & 196          &  -10.734     & 21          & -10.526   \\  \hline \hline
    \textbf{AVE.}                                  &  -          &   -                         &  -            &  \textbf{0.079}       &    -        & \textbf{0.999}     \\  \bottomrule
    \end{tabular}
    }
\end{table}

\begin{table}[t]
    \centering
    \caption{Comparison of the results of ``pre-mapping''.}
    \label{table:post-mapping}
    \setlength{\tabcolsep}{1pt}
    \resizebox{\linewidth}{!}{
    \begin{tabular}{c|ccc|cccccc}
    \toprule
    \multicolumn{1}{c|}{\multirow{3}{*}{Circuits}} & \multicolumn{3}{c|}{refactor+(fpga-mapping)}                & \multicolumn{6}{c}{refactor-plus+(fpga-mapping)}             \\
    \multicolumn{1}{c|}{}                          & edge          & area       & \multicolumn{1}{c|}{delay } & edge         &  improv\%    & area     &  improv\% & delay   &  improv\% \\ \hline \hline
    log2                                           & \textbf{36397}         & 52225      & 76                 & 36572        &  -0.481      & \textbf{50600}    &  3.112    & \textbf{73}      & 3.947   \\
    square                                         & 16518         & 34848      & 50                          & \textbf{16437}&  0.490       & \textbf{34451}    &  1.139    & 50      & 0       \\
    adder                                          & 1036          & 1666       & 51                          & \textbf{1022} &  1.351       & \textbf{1654}     &  0.720    & 51      & 0       \\
    sin                                            & 6491          & 9786       & 42                          & \textbf{6359} &  2.034       & \textbf{9146}     &  6.540    & \textbf{41}      & 2.381   \\
    div                                            & \textbf{76998}         & 84553      & 864                & 77016         &  -0.023      & \textbf{84298}    &  0.302    & 864     & 0       \\
    hyp                                            & 197758        & 342434     & 4194                        & \textbf{197728}&  0.015       & \textbf{342162}   &  0.079    & \textbf{4193}    & 0.024   \\
    max                                            & 3350          & 3365       & 56                          & \textbf{3141}  &  6.239       & \textbf{3130}     &  6.984    & \textbf{46}      & 17.857  \\
    sqrt                                           & \textbf{21874}         & 34280      & 1023               & 21900         &  -0.119      & \textbf{34116}    &  0.478    & \textbf{1022}    & 0.098   \\
    multiplier                                     & \textbf{27704}         & 45326      & 53                 & 27818         &  -0.411      & \textbf{43502}    &  4.024    & 53      & 0       \\
    bar                                            & 2688          & \textbf{4476}       & 4                  & 2688          &  0           & 4539              &  -1.408   & 4       & 0       \\
    priority                                       & \textbf{1004} & \textbf{1279}       & \textbf{30}                 & 1064          &  -5.976      & 1393              &  -8.913   & 31      & -3.333  \\
    cavlc                                          & 629           & 966        & 4                           & \textbf{619}  &  1.590       & \textbf{952}      &  1.449    & 4       & 0       \\
    arbiter                                        & 15137         & 12415      & 18                          & \textbf{15135}&  0.013       & 12415             &  0        & 18      & 0       \\
    i2c                                            & 1662          & 1565       & 4                           & \textbf{1613} &  2.948       & \textbf{1513}     &  3.323    & 4       & 0       \\
    voter                                          & 12493         & 38111      & \textbf{15}                          & \textbf{12308}&  1.481       & \textbf{32187}    &  15.544   & 16      & -6.667  \\
    int2float                                      & 250           & 289        & 3                           & \textbf{241}  &  3.600       & \textbf{276}      &  4.498    & 3       & 0       \\
    ctrl                                           & 133           & 232        & 2                           & \textbf{131}  &  1.504       & \textbf{194}      &  16.379   & 2       & 0       \\
    dec                                            & 684           & 397        & 2                           & 684           &  0           & 397               &  0        & 2       & 0       \\
    mem\_ctrl                                      & 54008         & 61254      & \textbf{25}                          & \textbf{51361}&  4.901       & \textbf{60039}    &  1.984    & 28      & -12      \\
    router                                         & 310           & \textbf{386}  & 11                       & \textbf{302}  &  2.581       & 401               &  -3.886   & \textbf{6}       & 45.455   \\  \hline \hline
    \textbf{AVE.}                                  &  -            &  -         &   -                         &  -           & \textbf{1.087}  &    -     &  \textbf{2.617}    & -       & \textbf{2.388}    \\  \bottomrule
    \end{tabular}
    }
\end{table}

The ``balance'' command in the berkeley-abc tool is a powerful technique for reducing the depth of AIG circuits.
Table \ref{table:post-balance} presents the results of applying the ``refactor'' and ``refactor-plus'' operators followed by the ``balance'' command.
 Despite the effectiveness of the ``balance'' operation, the ``refactor-plus'' still achieves an average improvement of 2.436\% in size and 0.939\% in depth compared to ``refactor''. 

\subsubsection{Comparison by ``resyn2''}
The ``resyn2'' is a logic optimization flow in the berkeley-abc tool, and it is composed of the sequence of ``\textit{balance, rewrite, refactor, balance, rewrite, rewrite -z, balance, refactor -z, rewrite -z, balance}''.
We use ``refactor-plus'' to replace the optimization of ``refactor'' and ``refactor -z''.
Most of the cases as shown in Table \ref{table:resyn2} show that the ``resyn2'' with refactor can get improvement, in general, we have average improvements for the ``resyn2'' with 0.079\% in size and 0.999\% in depth.

It should be noted that the ``resyn2'' is a heuristic approach, the integration of refactor-plus to ``resyn2'' is not necessarily the most appropriate.
However, this experiment result still shows that refactor-plus brings the difference of optimization space. 

\subsubsection{Comparison by ``pre-mapping''}
We also evaluated the optimization capabilities of ``refactor-plus'' from the perspective of technology mapping.
Table \ref{table:post-mapping} presents the results of applying the ``refactor'' and ``refactor-plus'' operators followed by the FPGA technology mapping command ``if -K 6'' in the berkeley-abc tool. 
It shows that the ``refactor-plus'' achieves an average improvement of 1.087\% in edge, 2.617\% in area, and 2.388\% in delay compared to ``refactor''.

Overall, these experiments demonstrate that ``refactor-plus'' is a powerful approach that effectively reduces the size and depth of the circuits, making it a valuable approach for Boolean network optimization.

\subsection{Discussion}
\label{sec:Eval-Discussion}
This paper introduces a reconvergence-driven AIG rewriting method called ``refactor-plus'' and demonstrates its effectiveness through a series of experiments.
The multi-node-rewriting algorithm can expand the searching space, while strategy learning can help to make a good decision for a given cone.

One of the notable strengths of ``refactor-plus'' is its ability to reduce both the size and depth of the optimized circuits.
The expanded search space leads to better possible structure candidates, and the \textit{is\_cirtical} and \textit{max\_depth} is also considered as part of features when training the classifier.
What this does is it allows ``refactor-plus'' to achieve an average improvement of 5.567\% in size and 5.327\% in depth.

The convergence of both the Q-learning-based RL and MLP-based classification procedures is essential for the overall effectiveness of the proposed method.
First, the problem of ``refactor-plus'' can be formulated as an MDP, theoretically, there is an optimal solution.
Therefore, we can get a good training dataset from the Q-learning-based RL to label each cone's feature.
With this good dataset, a good classification result is obtained with a test accuracy of 87\%.


\section{conclusion}
\label{sec:conclu}
This paper proposes a new approach to AIG rewriting that utilizes two key techniques: multi-strategy-based rewriting and strategy learning. 
The adaptive reconvergence-driven approach expands the search space for possible structures by allowing for a multi-node-rewriting algorithms selection, which is in contrast to the traditional method. 
The strategy learning process involves two parts: generating a training dataset through RL and then classifying it using MLP.

Our evaluation results demonstrate that this new approach significantly improves the QoR of reconvergence-driven AIG rewriting when compared to the traditional method. 
However, there are still many areas for exploration, such as finding the optimal operator sequence for circuits and applying for multi-outputs-window-based rewriting algorithms. 
In the future, we plan to continue exploring these areas and refining our approach.


\bibliographystyle{IEEEtran}
\bibliography{optmizer.bib}

\end{CJK*}
\end{document}

%% file: config.tex
\usepackage{CJKutf8}
\usepackage{changes}

\usepackage{gensymb}
\usepackage{bm}

\usepackage{amsmath,amssymb}
\usepackage{amsthm}

\usepackage{balance}

\usepackage[figuresright]{rotating}

\usepackage{graphicx}
\graphicspath{{figures/}}
\usepackage[export]{adjustbox}
\usepackage{flushend}

\usepackage{subfigure}
\usepackage{float}

\usepackage{listings}
\usepackage{color}
\usepackage{soul}
\usepackage{tikz}
\newcommand*{\circled}[1]{\lower.7ex\hbox{\tikz\draw (0pt, 0pt)%
    circle (.5em) node {\makebox[1em][c]{\small #1}};}}

\usepackage{algorithm}
\usepackage{algpseudocode}

\theoremstyle{definition}

\newtheorem{example}{Example}

\newtheorem*{proof}{Proof}
 
\theoremstyle{remark}

\definecolor{mygreen}{rgb}{0,0.6,0}
\definecolor{mygray}{rgb}{0.5,0.5,0.5}
\definecolor{mymauve}{rgb}{0.58,0,0.82}

\usepackage{booktabs} 
\usepackage{multirow}
\usepackage{array}
\newcolumntype{I}{!{\vrule width 1.2pt}}
\newlength\savedwidth

\newlength\savewidth

\usepackage{makecell}

\usepackage{etoolbox}
\makeatletter
\patchcmd{\@makecaption}
  {\scshape}
  {}
  {}
  {}
\makeatletter
\patchcmd{\@makecaption}
  {\\}
  {.\ }
  {}
  {}
\makeatother

\usepackage{geometry}